\documentclass[runningheads]{llncs}
\usepackage{cite}
\usepackage[ruled,vlined]{algorithm2e}
\SetKwInput{kwStep}{Step 1}
\SetKwInput{kwStepp}{Step 2}
\usepackage{multirow}
\usepackage{enumitem}
\usepackage{xcolor,colortbl}

\definecolor{Gray}{gray}{0.85}
\definecolor{LightCyan}{rgb}{0.88,1,1}

\usepackage{amsmath,amssymb,amsfonts}
\usepackage[ruled,vlined]{algorithm2e}
\usepackage{algorithmic}
\usepackage{graphicx}
\usepackage{textcomp}
\usepackage{booktabs}
\usepackage{color, colortbl}
\definecolor{Gray}{gray}{0.9}
\usepackage{bbm}
\usepackage{epsfig}
\usepackage[utf8]{inputenc}
\usepackage{float}
\usepackage{caption}
\usepackage{subcaption}

\begin{document}

\title{Incremental Semi-supervised Federated Learning for Health Inference via Mobile Sensing}

\author{Guimin~Dong
    \thanks{Guimin Dong is with Amazon.com (e-mail: guimind@amazon.com), }%
        Lihua~Cai
    \thanks{Lihua Cai is with South China Normal University, China (email: lee.cai@m.scnu.edu.cn),}%
        Mingyue~Tang,
        Laura~E.~Barnes,
        and~Mehdi~Boukhechba
    \thanks{Mingyue Tang, Laura E. Barnes,
        and Mehdi Boukhechba are with University of Virginia, Charlottesville, VA 22903 USA (email: utd8hj,lb3dp,mob3f@virginia.edu).}%
    \thanks{Guimin Dong and Lihua Cai contributed equally.}
    }
\institute{}

\maketitle

\begin{abstract}
Mobile sensing appears as a promising solution for health inference problem (e.g., influenza-like symptom recognition) by leveraging diverse smart sensors to capture fine-grained information about human behaviors and ambient contexts. Centralized training of machine learning models can place mobile users' sensitive information under privacy risks due to data breach and misexploitation. Federated Learning (FL) enables mobile devices to collaboratively learn global models without the exposure of local private data. However, there are challenges of on-device FL deployment using mobile sensing: 1) long-term and continuously collected mobile sensing data may exhibit domain shifts as sensing objects (e.g. humans) have varying behaviors as a result of internal and/or external stimulus; 2) model retraining using all available data may increase computation and memory burden; and 3) the sparsity of annotated crowd-sourced data causes supervised FL to lack robustness. In this work, we propose FedMobile, an incremental semi-supervised federated learning algorithm, to train models semi-supervisedly and incrementally in a decentralized online fashion. We evaluate FedMobile using a real-world mobile sensing dataset for influenza-like symptom recognition. Our empirical results show that FedMobile-trained models achieve the best results in comparison to the selected baseline methods.
\end{abstract}


\section{Introduction} \label{intro}

Mobile devices such as smartphones, smartwatches, and other wearable devices are equipped with a rich set of sensors that can collect human behavioral and physiological data continuously and unobtrusively \cite{guiminChil2021}. Data collected by using embedded sensors (e.g., accelerometer, GPS, and Bluetooth sensors) in mobile devices have been leveraged in plethora healthcare-related fields including but are not limited to physical state inference, mental health monitoring, and mobile interventions \cite{zhiyuanwang1, zhiyuanwang2}. However, wide adoption of mobile sensing applications, especially for health inference problems such as influenza-like symptom recognition, remains challenging due to three major concerns: 1) privacy issues, 2) long-term streaming data, and 3) sparse labeled data.

For privacy concerns, where the data are stored, who has access to them, and how they are used are among the typical aspects of users' concerns \cite{cui2023communicationefficient}. For example, GPS trajectories can reveal mobile users' whereabouts, which can then be exploited to further infer sensitive personal information such as race, gender, physical activities, social relationship, and health status. A second example involves adopting Bluetooth devices for contact tracing, as users who contract contagious viruses are generally required to release all their personal information. Other users may also have concerns about being tracked of their social relationships, as they may not know how their contact data might be used beyond monitoring infectious disease.

A privacy-preserving method is needed to address these privacy concerns in mobile sensing applications. Privacy preservation is both a legal and technical problem, and cannot be simply absolved by a voluntary consent form from users \cite{liuhan1,liuhan2}. Appropriate technical and organizational measures are required through a proper engineering process. Many privacy-preserving approaches and algorithms have been proposed according to different application scenarios. 
For instance, k-anonymity algorithm and its variants have been proposed to hide identities in the source of data items~\cite{guimindong1, Cui2022Federated}. 
Differential privacy aims to reduce exposure of authentic information by adding perturbations that follow certain noise distribution while maintain data utility. 
However, anonymization techniques could make sensitive attributes vulnerable to inference attacks; and differential privacy could downgrade data utility and lower the power of learned models \cite{yin2021comprehensive}, whereas model utilities are critical in mobile health applications. 
Federated learning (FL) is a decentralized learning mechanism in which users can collaboratively contribute knowledge to the global learning target without transmitting data samples to a central server, or sharing them among users \cite{yang2019federated}. 
In mobile sensing applications using FL, mobile devices hold their own datasets and use them to train local models, reducing the chance of data and privacy breach; a central server then trains a global model by learning from local weights or gradients. 

For the long-term streaming data problem, data collection of these applications is often a prolonged process. Take influenza-like symptom recognition as an example, only long-term monitoring can be meaningful and effective, as influenza symptoms usually exist for few days, while people will stay in good health for the majority of time. Short term monitoring cannot guarantee capturing participants' symptoms and their associated behavior changes when they are unwell. 
Meanwhile, if we retrain models using all existing data, the computation and memory burden would be high and prohibitive for mobile devices \cite{Wen2021Hardware, Wen2021FPGA, Wen2021Software}. If we retrain models using only the newly incoming data, a classic problem referred to as catastrophic forgetting will significantly downgrade model performance.
Incremental learning is designed for models built on streaming data to mitigate the catastrophic forgetting problem \cite{mai2021online} by continuously updating the models with knowledge transfer. It also helps the model continuously monitor users' health states, while adapt to computation and memory constraints in mobile devices \cite{Wen2021OpenMem,Wen2021Software}. 

Lastly, for the sparse labeled data problem, data annotations are often carried out by the users in mobile sensing applications. These tasks add significant burdens on participants and result in low compliance rate and high attrition rate. To mitigate user burdens, annotation tasks would be limited, leading to sparse labeled data \cite{chen2021uniting, chen2022explain,suiyaochen2}. Thus the robustness of the learned models built on datasets with limited and sparse labels cannot be guaranteed. 
A semi-supervised method could be incorporated to enhance the data with limited labels to ensure the utility and robustness of the learnt models \cite{luo2023efficient, luo2019Training, luo2020intemitten}.

Combining the aforementioned solutions for a privacy preserving mobile sensing framework, we employ graph neural network as the basic model as they have been proved to be capable of capturing hierarchical spatial and temporal relationships in human behavior dynamics (i.e., people's activities can vary over time and locations)~\cite{guiminChil2021}.  
We propose an incremental semi-supervised federated learning approach, FedMobile, in an attempt to address the above challenges. To demonstrate our approach, we applied it using an in-house proprietary mobile sensing dataset on an influenza symptom recognition task. Our contributions are summarized as follows:

\begin{itemize}
\item We propose an incremental federated learning (IFL) framework, FedMobile, to train neural network models in a decentralized approach by using Knowledge distillation, which can transfer expert knowledge to apprentices. This incremental framework is designed to update existing models only using newly available data in an online fashion.
\item We integrate our above approach with a semi-supervised method using Consistency Regularization (CR), which can utilize massive amount of unlabeled data to enhance model robustness.
\item Our proposed FedMobile is a general privacy-preserving method that is applicable for multitudinous mobile sensing and machine learning applications. The framework is proven effective in the health inference task, influenza-like symptom recognition, by using a real-world mobile sensing dataset that was collected in the wild.  
\end{itemize}

\begin{figure*}[t]
\begin{center}
\includegraphics[width=0.75\textwidth]{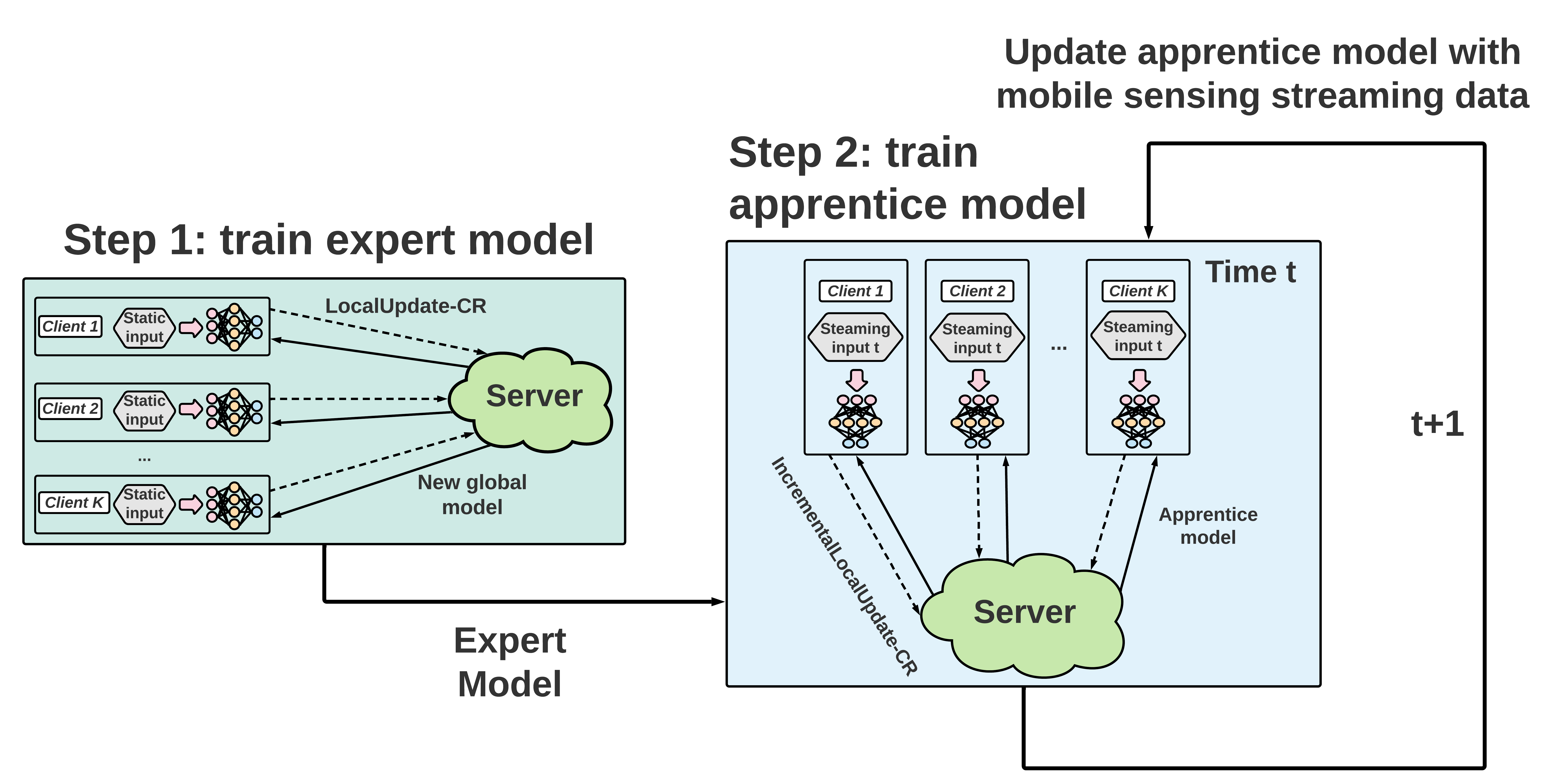}
\caption{The general framework of our FedMobile}
\label{fig:incrementalFL}
\vspace{-3mm}
\end{center}
\end{figure*}

\section{Related work} \label{related_work}

As more and more fine-grained personal data are being collected and applied in different health inference tasks, privacy protection has become an essential demand by the data owners, and federated learning as a general framework can fulfill a large degree of this demand by keeping individual data local while allowing development of artificial intelligence models in a distributed manner~\cite{yang2019federated,suiyaochen3}. Localizing personal data allows data owners to have full control over how their data can be accessed and used, reduces risks of unintended data exposures, and makes possible collaborative developments of machine learning models that require large scale training data.
It is also not in direct conflicts with many other privacy-preserving techniques such as homomorphic encryption~\cite{wood2020homomorphic,yifanmao3}, thus can be combined with them to further enhance privacy preservation.

Ever since its first proposal by a few Google engineers in an application to Android device keyboard stroke prediction~\cite{yang2018applied}, federated learning has been widely researched and applied in many fields (e.g., natural language processing (NLP), internet of things (IoT). Among all these researches in FL, of special interests to us are works that fall in mobile edge computing, health related inferences, and the intersection between them.
Empowered by 5G network, as we move towards the internet-of-things era, applications of FL in mobile edge computing are increasingly commonplace.
Bonawitz et al. proposed a scalable production system for federated learning on Android mobile devices, of which the system design is not limited to the Android operating system~\cite{bonawitz2019towards,yifanmao1,suiyaochen1}.
Lim et al. disscussed different aspects of implementing federated learning in a mobile edge network, pointing out various implementation challenges in FL such as security attacks, computation resources constraints, and particularly the unlabeled data issue~\cite{lim2020federated}.
Albaseer et al. built a FedSemL system on the network edge with considerations in limited resources, deadline constraints, a scarcity of labeled data and an abundance of unlabeled data samples in edge devices~\cite{albaseer2022semi,yifanmao2}.

To tackle the problem of lack of labels in clients, semi-supervised learning is integrated to the FL framework to leverage the large amount of unlabeled data. 
Zhong et al. proposed a semi-supervised federated learning method for heterogeneous devices, which adopts pseudo-labeling method as its semi-supervised learning strategy~\cite{zhong2022semi}. Specifically, it assumes that the central server has a small amount of labeled data to train an initial global model, which is then used to obtain pseudo-labels for local unlabeled samples that meet certain entropy conditions (i.e., below a selected entropy threshold). 
Presotto et al. proposed SS-FedCLAR, a FL framework that combines Federated Clustering and Semi-Supervised Learning to mitigate both the non-IID and data scarcity problems for Personalized Sensor-Based Human Activity Recognition~\cite{presotto2022federated}. Specifically for semi-supervised learning, active learning and label propagation were combined to train local models without labeled data in clients.
Zhao et al. proposed a multimodal and semi-supervised
federated learning framework that trains auto-encoders to extract
shared or correlated representations from different local data
modalities on clients~\cite{zhao2022multimodal, luo2021smarton}. First the server utilizes central data in different modalities to train a multimodal auto-encoder (e.g., deep canonically correlated auto-encoders); then the global auto-encoder is shared among clients with different data modality orientations for further updates; these local updates on the auto-encoders will then be aggregated in the server and used to encode an auxiliary multimodal labeled dataset, from which a classifier will be built and used by all clients. In this fashion, the traditional FL problem is converted into an auto-encoder based FL problem that can accommodate heterogeneous data modalities in clients.
Yang et al. proposed Federated Incremental Learning (FedIL), a novel and general framework that employs a siamese network for contrastive learning to ensure acquiring high quality pseudo-labels during training~\cite{yang2023fedil}. 
Compared with these works, our current work adopts consistency regularization for semi-supervised learning, enabling the use of large amount of unlabeled data in clients.

In addition to the lack of data labels in clients, most realistic FL applications are also inherently incremental. As new data are generated in client devices, they are fed to local training rounds to obtain local model updates that will be aggregated by the central server for the global model updates. The new data used in this process could contain unseen new classes or shifts in domain (i.e., changes in the underlying distribution for the data due to user behavior changes), dramatically downgrading the performance in the original model~\cite{mai2021online}. This so called catastrophic forgetting phenomenon needs to be addressed when implementing FL applications.
In \cite{zhu2022attention}, the authors proposed an online traffic classification model called Fed-SOINN, which adopted a self-organizing incremental neural network in a hierarchical federated learning framework. Their online incremental learning component is dedicated to solve the unseen new network traffic classes in the online traffic classification problem.
Likewise, Dong et al. proposed a new model called Global-Local Forgetting Compensation (GLFC) to tackle the federated class-incremental learning problem~\cite{dong2022federated}.
Yang et al. applied KL loss to enforce the consistency between the predictions made by clients and the server during client training, while screened the uploaded client weights by cosine similarity with normalization to accelerate the convergence of model training~\cite{yang2023fedil}. 
Castellon et al. proposed to address the domain shift challenge by clustering clients into groups with similar data distributions, effectively creating global models for each cluster of clients~\cite{castellon2022federated}.
In comparison, our proposed approach leverages the advantages of incremental learning (IL) to federally update model weights in streaming data by continuously transferring previous learned knowledge from previously trained models to new models.

In healthcare domain, federated learning is also increasingly explored by researchers with data in the format of electronic health records (EHRs) and Internet-of-Medical-Things (IoMT) sensor data.
Vaid et al. predicted mortality in hospitalized COVID-19 patients within seven days by logistic regression and Multi-Layer Perceptron (MLP) federated models \cite{vaid2021federated}.
Wang et al. proposed a Privacy Protection Scheme for Federated Learning under Edge Computing (PPFLEC), which consists of a lightweight privacy protection protocol, and an algorithm based on a digital signature and hash function~\cite{wang2022privacy}.
Ahmed et al. proposed a structural hypergraph as well as an emotional lexicon for word representation using attention-based mechanism and federated learning for mental health symptom detection~\cite{ahmed2022hyper}.
Salim et al. proposed a Federated Learning-based Electronic Health Record sharing scheme for Medical Informatics to preserve patient data privacy~\cite{salim2022federated}.
In our work, we apply federated learning in solving a practically challenging recognition task on influenza symptoms using mobile sensing data from over 800 participants.

\section{Proposed Method}
In this section we introduce our proposed FedMobile framework. In general, there are two steps in FebMobile, as shown in Fig \ref{fig:incrementalFL}: step 1) train an expert model decentralized and semi-supervisedly by using federated learning with consistency regularization; step 2) train an apprentice model and continuously update the model parameters with the newly incoming batch of mobile sensing data using the same semi-supervised federated learning (semiFL) framework as the first step. In the training process of apprentice models and continual update of the model parameters, we integrate semiFL with knowledge distillation to transfer previously learned knowledge to the current model without retraining the model from scratch, in order to alleviate device memory burden and mitigate overfitting caused by newly collected data from various domains.

\subsection{Step 1: training an expert model decentralizedly using consistency regularized local update.} \label{section:step1}

Consistency regularization (CR) is a widely-used technique for semi-supervised and self-supervised learning, aiming to improve the generalizability of a trained model by leveraging a large amount of unlabeled data \cite{mustafa2020transformation}. The core idea of CR is to penalize the models that are sensitive to perturbed inputs under the assumption that the label semantics are not affected by perturbations. In FL setting, especially in practice with non-IID data, FL algorithm can encounter a weight divergence problem as the global model goes through increasing rounds of communications. To utilize the large amount of unlabeled data, we use CR as regularizer by comparing model outputs of original unlabeled data and unlabeled data with perturbations. Here we assume that, for each round of communications, the mini batches of unlabeled data within each client remain consistent for small perturbations. We introduce a new loss component, Kullback–Leibler divergence, $KL(\bullet||\bullet)$, to calculate the divergence between the prediction output distributions of the unlabel data and the unlabel data with perturbations. This will penalize the local models if they try to update their model weights in significantly divergent and locally overfitting directions. The loss function of local update with inter-communication CR, for epoch $i$ and batch $j$, is expressed as 
\begin{equation}
\begin{split}
    \mathcal{L}(\mathbf{W}) &= \mathcal{L}_{CE}(\mathbf{\hat{y}}_{i,j}, \mathbf{y}_{i,j}) \\ &+ \mathcal{L}_{CR}( \mathcal{F}(\sigma_{i,j}),\mathcal{F}(\sigma^{'}_{i,j})) + \mathcal{R}(\mathbf{W}),
\end{split}
\end{equation}
where $\mathcal{L}_{CE}(\bullet)$ is a cross entropy loss function, $\sigma_{i,j}$ and $\sigma^{'}_{i,j}$ are the logit outputs generated from current DL model using the original unlabeled data and perturbed unlabeled data  respectively. $\mathcal{L}_{CR}(\bullet)$ is the consistency regularization loss, where $\mathcal{L}_{CR} = \lambda KL(\mathcal{F}(\sigma_{i,j})||\mathcal{F}(\sigma^{'}_{i,j}))$ and $\mathcal{F}(\bullet)$ is a softmax function, $\lambda \in (0,1)$ is the CR coefficient. $\mathcal{R}(\bullet)$ is the regularization function (e.g. L2-norm). We present $\textbf{LocalUpdate-CR}$ in Algorithm \ref{alg:localupdate}. During Step 1 of training expert model distributedly in Figure \ref{fig:incrementalFL}, \textbf{LocalUpdate-CR} leverages both a small amount of labeled data and a large amount of unlabeled data to update the weights of client models using federated learning with CR, and then uploads the updated weights to the server to update the global model. Finally, at the end of step 1, an expert model is fed into the next stage of incremental FL to train an apprentice model using streaming data without retraining the model from scratch. 

\begin{algorithm} 
\SetAlgoLined
\SetKwFunction{Flocalupdate}{($k, \mathbf{W}, \mathcal{M}, \mathbf{X}, \mathbf{X}^{u}, \mathbf{y}$)}
\SetKwProg{Fn}{LocalUpdate-CR}{:}{}
\Fn{\Flocalupdate}{
        \For{each epoch $i = 1, ..., E$}{
        \For{each batch $j = 1, ..., B$}{
        $\mathbf{\hat{y}}_{i,j}\leftarrow \mathcal{M}(\mathbf{X}_{i,j})$\;
        $\sigma_{i,j}, \sigma^{'}_{i,j} \leftarrow \mathcal{M}(\mathbf{X}_{i,j}^{u}), \mathcal{M}(d(\mathbf{X}_{i,j}^{u}))$\;
        $\mathcal{L}(\mathbf{W}) \leftarrow 
       \mathcal{L}_{CE}(\mathbf{\hat{y}}_{i,j},\mathbf{y}_{i,j}) + \mathcal{L}_{CR}(\mathcal{F}(\sigma_{i,j}),\mathcal{F}(\sigma^{'}_{i,j})) + \mathcal{R}(\mathbf{W})$\;
        $\mathbf{W} \leftarrow \mathbf{W} - \eta \nabla \mathcal{L}(\mathbf{W}) $\;
        }
        }
        \Return {$\mathbf{W}$}\;
  }
\caption{Local update with consistency regularization. During each local update for client $k$, we calculate the inter-communication consistency regularized loss, where overfitting weight updates can be penalized and the weight divergence problem can be mitigated. $\mathbf{W}$ is the DL model weight, $\mathcal{M}$ is the DL model, $\mathbf{X}$ and $\mathbf{X^u}$ are labeled and unlabeled input data, $d(\bullet)$ is a stochastic perturbation function (e.g., Gaussian noise)\cite{chenwang1,chenwang2}, and $\eta$ is the learning rate.}
\label{alg:localupdate}
\end{algorithm}

\begin{figure*}[t]
\begin{center}
\includegraphics[width=0.75\textwidth]{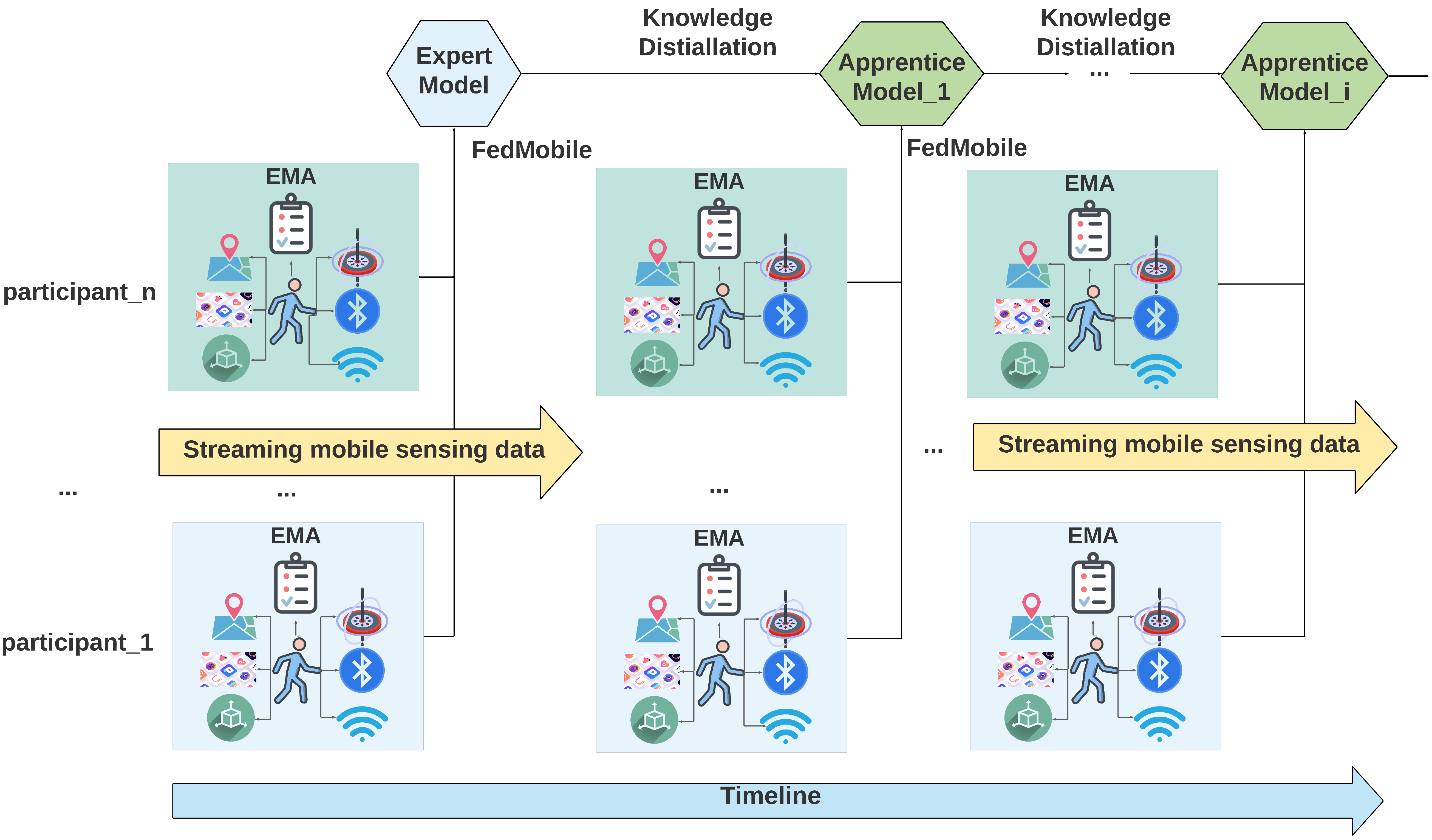}
\caption{FedMobile for streaming mobile sensing data.}
\label{fig:mobilesensing}
\vspace{-3mm}
\end{center}
\end{figure*}

\subsection{Step 2: training apprentice model and continual model learning using incremental local update.} \label{section:step2}

In general, knowledge distillation (KD) consists of two components, training expert model and training apprentice model. The expert model refers to a model trained by using historical data from domain population or a pre-trained model. In the distillation process for training apprentice model, knowledge learned by the expert model is transferred to the apprentice model by minimizing the distance of the prediction distributions between them given the same inputs \cite{ hinton2015distilling}. In the continual model learning process, KD allows the trained apprentice model to update its parameters by using the newly collected data, therefore historical data can be removed to alleviate memory burden in devices. Additionally, in the long run of mobile sensing data collection, the sensing objects can change their behaviors because of internal and/or external stimuli. Dong et al. \cite{dong2021detection}, for example, observe that during COVID-19, people become more sedentary and spend more time at home due to infection risk reduction and public intervention. Therefore, the collected data can come from varying domains, causing domain shift problems that downgrade the generalizability of previously trained models \cite{ayranci2022distinguishing}. KD enables the apprentice model to adapt to the newly collected data and overcome knowledge forgetting. In incremental FL, given the incoming batch of streaming data at time $t$, which includes both labeled and unlabeled data, the loss function for the client local update at epoch $i$ and batch $j$, is expressed as:
\begin{equation}
\begin{split}
    \mathcal{L}(\mathbf{W}^{\rho}) &= (1-\alpha)\mathcal{L}_{CE}(\mathbf{\hat{y}}_{t,i,j}, \mathbf{y}_{t,i,j}) \\ &+
    \alpha \mathcal{L}_{KD}(\mathcal{F}(\zeta^{\rho}_{t,i,j}),\mathcal{F}(\zeta^{\tau}_{t,i,j})) \\
     &+\mathcal{L}_{CR}(\mathcal{F}(\sigma_{t,i,j}),\mathcal{F}(\sigma^{'}_{t,i,j}))+\mathcal{R}(\mathbf{W}^{\rho}),
\end{split}
\end{equation}
where $\mathbf{W}^{\rho}$ is the apprentice model weights, $\alpha \in (0,1)$ is a balancing weight, $\zeta^{\rho}$ and $\zeta^{\tau}$ are the logit outputs of labeled data from the expert model at time 0 or apprentice model at time $t-1$ and apprentice model (at $t$) respectively. $\mathcal{L}_{KD}$ is the knowledge distillation loss, and 
$\mathcal{L}_{KD}(\mathcal{F}(\zeta^{\rho}_{t,i,j}),\mathcal{F}(\zeta^{\tau}_{t,i,j}))=KL(\mathcal{F}(\zeta^{\rho}_{t,i,j})||\mathcal{F}(\zeta^{\tau}_{t,i,j}))$ \cite{hinton2015distilling, gou2021knowledge}. 
\textbf{IncrementalLocalUpdate-CR} is presented in Algorithm \ref{alg:increlocalupdate}. In Step 2 shown in Figure~\ref{fig:incrementalFL}, given streaming data input at time $t$, incremental FL uses \textbf{IncrementalLocalUpdate-CR} to continually update client local models and then sends the updated weights to the server, which further aggregates the client model weights and updates the global model weights. 

\begin{algorithm} 
\SetAlgoLined
\SetKwFunction{Finlocalupdate}{($k, \mathbf{W}^{\rho}, \mathcal{M}^{\rho}, \mathcal{M}^{\tau}, \mathbf{X}_t, \mathbf{X}^{u}_t, \mathbf{y}_t$)}
\SetKwProg{Fn}{IncrementalLocalUpdate-CR}{:}{}
\Fn{\Finlocalupdate}{
        \For{each epoch $i = 1, ..., E$}{
        \For{each batch $j = 1, ..., B$}{
        $\hat{\mathbf{y}}_{t,i,j}, \zeta^{\rho}_{t,i,j} \leftarrow \mathcal{M}^{\rho}(\mathbf{X}_{t,i,j})$\;
        $\zeta^{\tau}_{t,i,j} \leftarrow \mathcal{M}^\tau(\mathbf{X}_{t,i,j})$\;
        $\sigma_{t,i,j}, \sigma^{'}_{t,i,j} \leftarrow \mathcal{M}^{\rho}(\mathbf{X}_{t,i,j}^u), \mathcal{M}^{\rho}(d(\mathbf{X}_{t,i,j}^u))$\;
        $\mathcal{L}(\mathbf{W}^{\rho}) \leftarrow (1-\alpha)\mathcal{L}_{CE}(\mathbf{\hat{y}}_{t,i,j}, \mathbf{y}_{t,i,j}) + \alpha\mathcal{L}_{KD}(\mathcal{F}(\zeta^{\rho}_{t,i,j}),\mathcal{F}(\zeta^{\tau}_{t,i,j}))+\mathcal{L}_{CR}(\mathcal{F}(\sigma_{t,i,j}),\mathcal{F}(\sigma^{'}_{t,i,j}))+\mathcal{R}(\mathbf{W}^{\rho})$\;
        $\mathbf{W}^{\rho} \leftarrow \mathbf{W}^{\rho} - \eta \nabla \mathcal{L}(\mathbf{W}^{\rho}) $\;
        }
        }
        \Return {$\mathbf{W}^{\rho}$}\;
  }
\caption{Incremental local update with consistency regularization. In each incremental update for each batch of incoming streaming data at time $t$, the apprentice model is updated locally by leveraging the knowledge generalized from the expert model using knowledge distillation from incremental learning. $\mathbf{W}^{\rho}$ is the model weight of the apprentice model, $d(\bullet)$ is a stochastic perturbation function (e.g., Gaussian noise), $\mathcal{M}^{\rho}$ and $\mathcal{M}^\tau$ are apprentice DL model and expert DL model respectively.}
\label{alg:increlocalupdate}
\end{algorithm}

\subsection{End-to-End Incremental Semi-supervised FL}
Based on \textbf{LocalUpdate-CR} and \textbf{IncrementalLocalUpdate-CR}, we present the end-to-end incremental FL with consistency regularization, as shown in Algorithm \ref{alg:incrementalfl}. 
Given the labeled input $X^{l}$, the corresponding target $y^{l}$ and unlabeled input $X^{u}$ from each client $k$, the goal of step 1 is to train a global expert model using semi-supervised FL with both labeled and unlabeled local data. 
First, the server initializes $W_0$ for the global model $\mathcal{M}_0$. Then for each round of communications, we run \textbf{LocalUpdate-CR} for each client $k$ to update the local models, and send the updated weights to the server. The server aggregates the new local weight updates using FedAvg
\cite{mcmahan2017communication}. After local weight aggregation, the global model weights are updated in the server. Finally, after $C$ rounds of communications, the first stage of incremental FL outputs the expert model denoted as $\mathcal{M}^{\tau}$.
In step 2, given the expert model $\mathcal{M}^{\tau}$, streaming inputs $X^{l}_{t,k}$, $X^{u}_{t,k}$ for each client $k$ at time $t$, we train an apprentice model $\mathcal{M}^{\rho}$ and continual update $\mathcal{M}^{\rho}$ in a decentralized manner using streaming data and adapt the knowledge learned from the expert model to the newly collected data. At each incremental iteration $t$, we run $C$ rounds of communications to train the apprentice model. In each round of communications, we run \textbf{IncrementalLocalUpdate-CR} to update local model weights and send the updated weights to the server.   

\begin{algorithm} 
\SetAlgoLined
\kwStep{train expert model}
\SetKwFunction{FIncrementalFL}{}
\SetKwProg{Fn}{Server executes}{:}{}
\Fn{\FIncrementalFL}{
        initialize $\mathbf{W}^{\tau}_0$ for global model $\mathcal{M}^{\tau}$\;
        \For{each round of communications $c = 0,1,2..., C$}{
        \For{each client $k \in K$ in parallel}{
        $\mathbf{W}^{\tau}_{c,k}\leftarrow$ LocalUpdate-CR($k, \mathbf{W}^{\tau}_{c-1,k}, \mathcal{M}^{\tau}, \mathbf{X}^{l}_{c,k}, \mathbf{X}^{u}_{c,k}, y_{c,k}$)\;}
        $\mathbf{W}^{\tau}_{c} \leftarrow \sum_{k=1}^K \frac{n_k}{n} \mathbf{W}^{\tau}_{c,k}$\;
        $\mathcal{M}^{\tau}$ $\leftarrow$ update $\mathcal{M}^{\tau}$ by $\mathbf{W}^{\tau}_c$\;
        }
        
        \KwRet global model $\mathcal{M}^{\tau}$ as expert model\;
  }
\kwStepp{train the apprentice model and continually update the model using streaming data and knowledge distilled from the expert model or previous apprentice model}
\SetKwFunction{FIncrementalFLs}{}
\SetKwProg{Fn}{Server executes}{:}{}
\Fn{\FIncrementalFLs}{
        initialize $\mathbf{W}^{\rho}_0$ for $\mathcal{M}^{\rho}$\;
        \For{streaming data at time $t = 0,1,2...$}{
        \For{each round of communications $c = 0,1,2..., C$}{
        \For{each client $k \in K$ in parallel}{
         $\mathbf{W}^{\rho}_{t,c,k}, \mathcal{M}^{\tau}\leftarrow$ IncrementalLocalUpdate-CR($k, \mathbf{W}^{\rho}_{t-1,c,k}, \mathcal{M}^{\rho},\mathcal{M}^{\tau}, \mathbf{X}^{l}_{t,c,k}, \mathbf{X}^{u}_{t,c,k}, 
         \mathbf{y}_{t,c,k})$;
        }
        $\mathbf{W}^{\rho}_{t,c} \leftarrow \sum_{k=1}^K \frac{n_k}{n} \mathbf{W}^{\rho}_{t,c,k}$\;
         $\mathcal{M}^{\rho}$ $\leftarrow$ update $\mathcal{M}^{\rho}$ by $\mathbf{W}^{\rho}_{t,c}$\;
        }
        $\mathcal{M}^{\tau} \leftarrow \mathcal{M}^{\rho}$
        }
        \KwRet apprentice model $\mathcal{M}^{\rho}$\;
  }
\caption{End-to-end incremental federated learning. As shown in  \ref{fig:incrementalFL}, this incremental FL method first trains a global expert model using FedAvg with \textbf{LocalUpdate-CR}, and then feeds the trained expert model into the second step of incremental FL. In the second step, a global apprentice model is trained using FedAvg with \textbf{IncrementalLocalUpdate-CR}.}
\label{alg:incrementalfl}
\end{algorithm}

\subsection{Graph Modeling in Mobile Sensing}
Handcrafted feature engineering has been widely investigated in machine learning studies of mobile sensing. Despite fine-grained featurization of raw mobile sensing data and sufficient interpretability, complex interactions and high-level interdependence within human state variations can be underrepresented in manually extracted features. In addition, there exists structural topology in non-euclidean domains of raw mobile sensing data (e.g., GPS, Bluetooth Encounter), it can be challenging for handcrafted feature engineering to capture such topological information \cite{dong2021semi, tang23b}. There are numerous study have been finished in investigating Graph Neural Networks (GNNs) for sensor data, and demonstrated advanced performance in \cite{dong2021detection, tang2022using}. We hypothesize that people who are in early stage of influenza will subconsciously become less active, and avoid unnecessary traveling and person-to-person interactions. We leverage multiple embedded sensors to capture human behavior information: GPS sensors are used to capture human mobility behaviors; Bluetooth encounters are used as proximity to measure social behaviors; accelerometer and gyroscope can monitor physical activeness; web-virtual behaviors are tracked by recording app usages; and ambient contexts can be inferred by WiFi signals \cite{guimindong1}.

\begin{figure}[t]
\includegraphics[width=0.8\columnwidth]{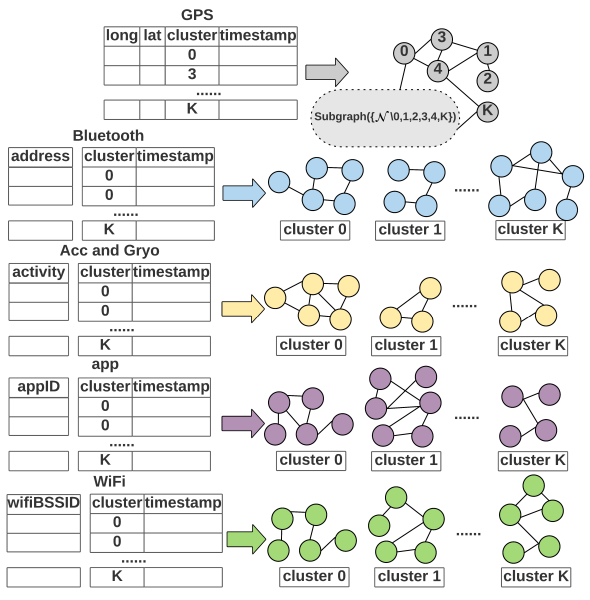}
\caption{ Graph Representations of Mobile Sensing Data.}
\label{fig:graph_ms}
\end{figure}

In graph construction process as shown in \ref{fig:graph_ms}, for mobility behavior, given a sequence of cluster labels generated from clustering of GPS coordinates that record people’s daily traveling trajectory, we treat each cluster label as a node and add edges between every pair of consecutive visited places (clusters). For the other behaviors and surrounding environments, we first partition the multi-modal sequential discrete sensor signals into several sub-sequences, where each sub-sequence represents the behaviors or contexts at the corresponding location people visit. For example, for accelerometer and gyroscope, we represent each recognized human activity as a node in graphs, and add edges between every two sequential activity states in the table. And then we apply this graph construction mechanism for each cluster partition to generate a sequence of graphs. Each sample in the input of GNNs consists of a static graph that encodes mobility behavior and multi-channel dynamic graphs that capture social, activity, web-virtual behaviors and ambient environments. Finally, we use the engineered graph structured data for early detection of influenza-like symptoms.

\begin{table*}[t]
    \centering
    \renewcommand{\arraystretch}{1.08}
  \resizebox{1\textwidth}{!}{
    \begin{tabular}{c|c|c|c|c|c|c|c|c}
    \toprule
    \multirow{2}{*}{Model}  &\multicolumn{4}{c|}{F1-Score} & \multicolumn{4}{c}{PR-AUC}\\
    \cline{2-9}
    
     & MLP & GCN & GAT &GCN-LSTM& MLP & GCN & GAT &GCN-LSTM\\
     \midrule
     Centralized & $0.733 \pm 0.010 $ & $0.751 \pm 0.007$ & $0.772 \pm 0.008$ & $0.811\pm 0.012$& $0.766 \pm 0.012$ & $0.795 \pm 0.011$ & $0.847 \pm 0.010$ & $0.882\pm 0.010 $
\\
    \midrule
     FedAvg &  $0.656 \pm 0.015$  & $0.663\pm0.013$ &$0.681\pm0.016$ & $0.724\pm 0.014$ &$0.687\pm0.009$ &$0.715\pm 0.012$&$0.739\pm0.009$ & $0.754\pm 0.011$
\\
  \midrule
    FedSem-FT & $0.675\pm 0.016$ & $0.698\pm0.015$ & $0.704\pm0.011$ &$0.736\pm 0.011$ &$0.706\pm 0.011$ &$0.724\pm0.013$ & $0.763\pm 0.012$ & $0.811\pm 0.015$ 
\\
    \midrule
    FedMatch-FT & $0.695\pm0.018$ & $0.711\pm0.014$  & $0.723\pm0.017$ & $0.757\pm 0.019$& $0.716\pm0.015$ & $0.753\pm0.014$ & $0.804\pm0.016 $ & $0.814\pm 0.012$
\\
\midrule
    FLwF & $0.703 \pm 0.019$ & $0.727 \pm 0.016 $ & $0.743 \pm  0.016$ & $0.781\pm 0.013$& $0.786 \pm 0.014$ & $0.715 \pm 0.017$ & $0.776 \pm  0.015$&$0.831\pm 0.008$
\\
    \midrule
FedMobile & $\mathbf{0.726 \pm 0.011}$ & $\mathbf{0.735 \pm 0.014}$ & $\mathbf{0.768 \pm 0.013}$  & $\mathbf{0.804\pm 0.015}$ &$\mathbf{0.742 \pm 0.010}$ &$\mathbf{0.788 \pm 0.012}$ &$ \mathbf{0.834 \pm 0.011} $&$\mathbf{0.861\pm 0.016}$
\\
  \bottomrule
\end{tabular}
}
\caption{Performance comparison between the selected baseline models and FedMobile.}
\label{table:result}
\end{table*}

\section{Data Collection and Experiment}
\subsection{Data Collection and Description}
We evaluate our proposed FedMobile by using real-world collected mobile sensing data that were collected to monitor warfighters' health status continuously. This mobile sensing study originates from a DARPA-funded project using smartphones for healthcare analysis, investigating passively collected mobile sensing data for earlier infectious disease diagnosis. The study protocol was approved by the University of Virginia IRB (IRB \#20770). Overall, 2,600 participants from 24 states in the U.S. signed informed consents and were recruited to participate in the data collection. The total length of this study was one year. A rich set of sensor data was collected, including GPS, Bluetooth encounter, Wifi signal, app usage, accelerometer, gyroscope, etc. We asked participants to install and run a mobile sensing app called ReadiSens. At 8 pm daily, ecological momentary assessments (EMAs) were delivered to collect self-reported influenza symptoms, including fever, cough, difficulty breathing, fatigue, muscle aches, headache, sore throat, runny nose, nausea, and diarrhea. We have the data from around 800 participants for this study. Active data collection was conducted on a rolling basis from February 15th, 2019 to April 30th, 2020. The total number of annotated and unannotated daily observations was around 6,100 (18\% were reported as having influenza symptoms) and 22,000, respectively. 

\subsection{Implementation details}
In the centralized training process and baseline Federated Learning methods, we divide the labeled data set into 70\%, 20\%, and 10\% subsets for training, validation, and testing, respectively. We adapt semi-supervised learning for the centralized training process. In the incremental FL training process, we simulate the mobile sensing streaming data and conduct FedMobile using the data in their chronological order. The first 40\% training data are used to conduct step 1 of FedMobile to produce an expert model by 40 rounds of communications as described in Section \ref{section:step1}. In the step 2 of FedMobile, as described in Section \ref{section:step2}, we set the number of incremental updates as 8 in this experiment, where the simulated streaming data were partitioned into 8 batches by their chronological order for the incremental training. The overall training process of FedMobile for streaming mobile sensing data is shown in Fig \ref{fig:mobilesensing}. And we set the number of rounds of communications as 10 for each batch of mobile sensing data empirically. Without considering the privacy issue, we train the models by a central server using Adam to investigate whether federated (privacy protected) methods can achieve comparable results with the centralized method. In the modeling of mobile sensing data for influenza-like symptom recognition, by leveraging graph representation of human behaviors as demonstrated in this related works \cite{dong2021detection,dong2021semi,cui2022nappn}, we transform raw mobile sensing data to graph structured data to extract high-level human behavioral features \cite{dong2021using}. We use Graph Convolutional Network (CCN) \cite{li2022scalable}, Graph Attention Network (GAT) \cite{tang2022using} and GCN-LSTM \cite{wu2021traffic} to fit the graph structured data, and use Multi-Layer Perceptron (MLP) \cite{cui2021geometric} to fit handcrafted features. We add Gaussian noise to the handcrafted features and randomly add/remove edges of graphs as the stochastic functions in consistency regularization for the semi-supervised FL. We set the default parameters across all models as batch size = 256, all embedding size = 128, and layer number=2. We train the models for 150 epochs. We optimize all models with AdamW optimizer. Xavier initialization is used to initialize the parameters. Learning rate is searched in [1e-6, 1e-5, 1e-4, 1e-3], and L2 normalization coefficient is tuned in [1e-7, 1e-6, 1e-5, 1e-4] using 30\% randomly sampled training dataset for all models.

\subsection{Baselines and Evaluation Metrics} \label{section:baseline}

a) Centralized Training (Centralized) is used as a baseline model to validate how the prediction power can be compromised by using FL approach. 

\noindent b) FedAvg \cite{mcmahan2017communication} is one of the classic supervised federated learning algorithms, and usually serves as baseline model in federated learning studies.

\noindent c) FedSem with Fine-Tuning (FedSem-FT) \cite{albaseer2020exploiting} is a semi-supervised federated learning method, which uses pseudo labeling for unlabeled data to improve the federated learning performance. To adapt FedSem with mobile sensing streaming data, we use Fine-Tuning to continually update the implemented neural networks. 

\noindent d) Federated Matching with Fine-Tuning (FedMatch-FT)  \cite{jeong2021federated} leverages a new inter-client consistency loss and disjoint learning for labeled and unlabeled data to perform federated learning semi-supervisedly. We used Fine-Tuning for FedMatch to continually update the neural networks given the streaming mobile sensing data.

\noindent e) Federated Learning without Forgetting (FLwF) \cite{usmanova2021distillation} is a distillation based FL approach that aims to tackle catastrophic forgetting in the class-incremental learning scenario.

Evaluation metrics: To evaluate model performance, we select F1-score and Precision-Recall AUC as our evaluation metrics: F1-score $ = \frac{2\cdot Prec \cdot Rec}{Prec + Rec}$ is a classification metric which balances precision $(Prec)  = TP/(TP+FP)$ and recall rate $(Rec) = TP/(TP+FN)$, where $TP$, $TN$, $FP$,  $FN$ are the number of true positives \cite{jin2021estimating}, true negatives, false positives, and false negatives; we calculate the area under precision-recall curves as PR-AUC. 

\begin{center}
\begin{figure*}[t]
\includegraphics[width=0.95\columnwidth]{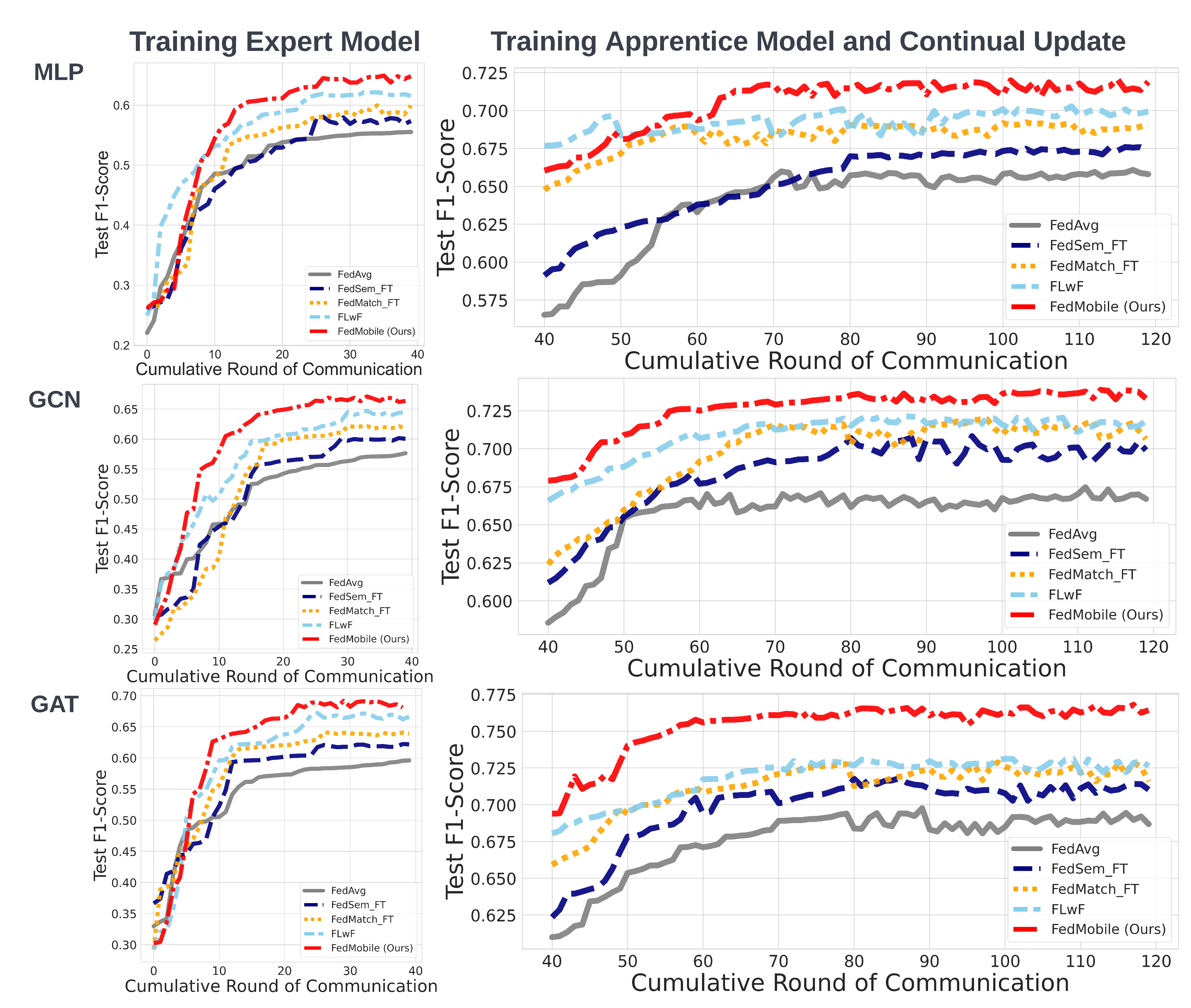}
\caption{Testing F1-Score comparison FedMobile and the baseline models for 120 communication rounds (epochs for centralized training). In the FedMobile, we train the expert models in the first 40 communication rounds and perform incremental update through 40-120 communication rounds.}
\label{fig:ifl_result}
\end{figure*}
\end{center}

\section{Results and Analysis}
\subsection{Performance Comparison}

The experimental results are shown in Table \ref{table:result}.  We observe that the models trained by using FedMobile have comparable results to the centralized models. As centralized training provides the best results without considering privacy issue, our proposed FedMobile shows equivalent prediction results, which indicates that our approach could achieve privacy preservation at the data level without compromising its prediction performance. FedMobile-trained models outperform the models trained by using the other FL baselines. As shown in Fig \ref{fig:ifl_result}, we observe that, generally, in the training of expert model, FedMobile shows a similar training trajectory with FedMatch-FT or FLwF at the begining and then surpasses the training trajectories of other baseline models; in the second step, FedMobile shows an continuously increasing F1-Score, which implies that FedMobile enables the trained model to gain incremental knowledge from both labeled and unlabeled data. 

The superior performance of FedMobile can be attributed to several components in the design of the incremental federated learning architecture. Firstly, the property of knowledge distillation (KD) imposes penalty for new knowledge to mitigate overfitting, and can adapt to the new knowledge after several iterations of incremental learning. Secondly, consistency regularization in FedMobile utilizes massive amount of unannotated samples to increase the robustness of GNNs by penalizing substantial changes of inferred label semantics between communication rounds \cite{chen2022grease}. Additionally, CR also reduces the divergence of the local weight updates among distributed clients to stabilize the global model parameter optimization.

\begin{center}
\begin{figure}[t]
\includegraphics[width=0.95\columnwidth]{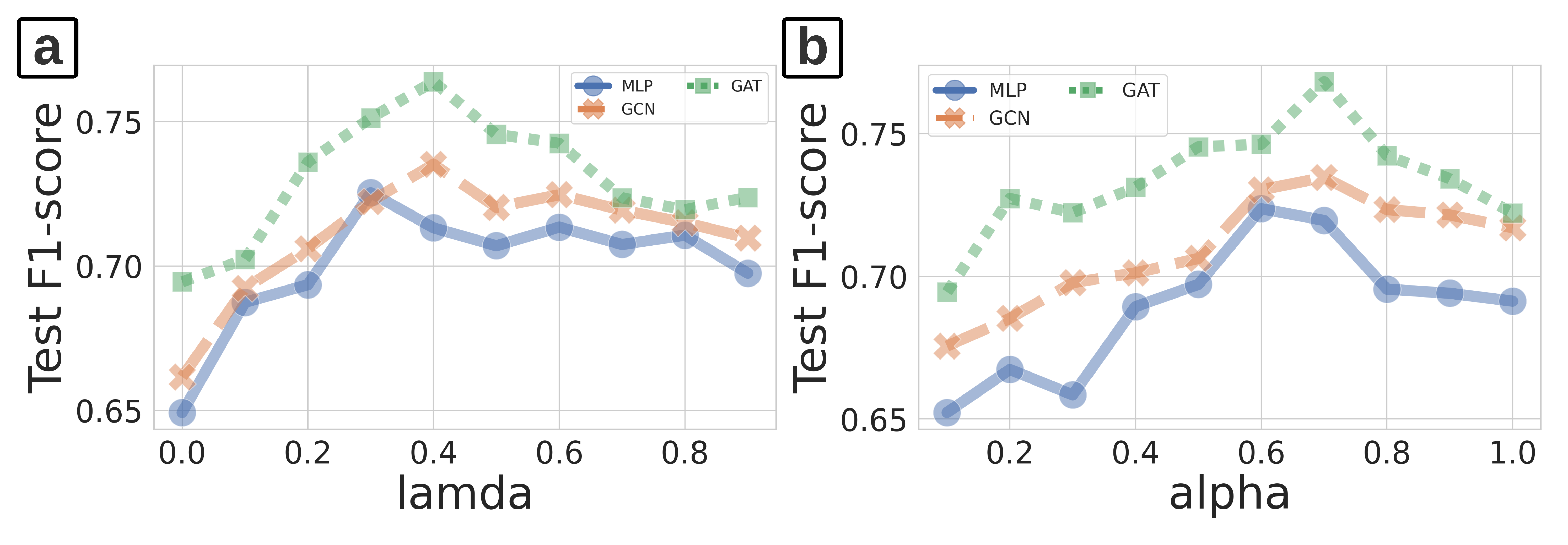}
\caption{Impact of (a) Consistency Regularization coefficient $\lambda$ and (b) KD loss coefficient $\alpha$ to the performance of FedMobile-trained model.}
\label{fig:ifl_hyperp}
\end{figure}
\end{center}

\begin{center}
\begin{figure}[t]
\includegraphics[width=0.95\columnwidth]{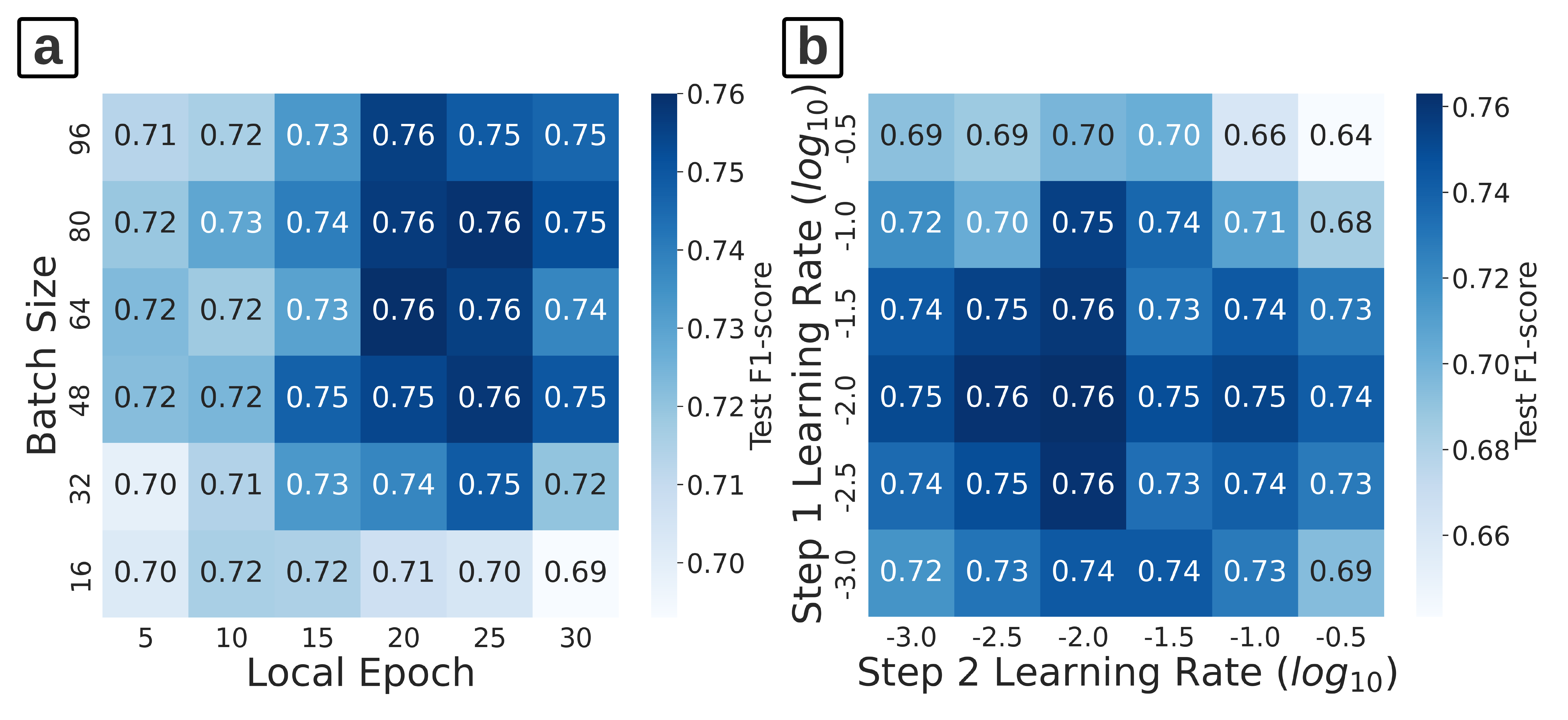}
\caption{The impact of (a) batch size and number of local epochs, and (b) learning rate in step 1 and 2 to FedMobile performance}
\label{fig:bsle}
\end{figure}
\end{center}
\subsection{Hyperparameter Analysis}

We analyze the impacts of CR coefficient $\lambda$ and the KD coefficient $\alpha$ on the performance of FedMobile-trained models. As shown in Figure \ref{fig:ifl_hyperp} (a), CR coefficient being 0.3/0.4 leads to the best performed models. This observation implies that making too much/little punishment (large/small $\lambda$) for the prediction results by using unlabeled data cannot mitigate within-client inconsistency, which results in less robust models. In the case when CR coefficient equals 0 (FedMobile without CR), downgrading performance can also imply the regularization effect of CR. 

In Figure \ref{fig:ifl_hyperp} (b), the models show less competitive performance when $\alpha$ is near 0 or 1, implying that complete reliance on either the expert model or the apprentice models can negatively affect model generalizability. 
And the value of $\alpha$ being 0.6/0.7 generates the best performance. This shows that previous learned knowledge can serve as regularization to prevent the learned models from overfitting by new incoming data with potential domain shifts.

In Figure \ref{fig:bsle} (a), we show the effects of combinations of local batch size and number of local epochs to FedMobile trained GAT. We observe that small batch size and small number of local epochs degrade GAT's prediction performance. Small batch size could make the predicted soft labels less representative and fail to reflect robust label distribution, and the CR less effective. 

In Figure \ref{fig:bsle} (b), we show that the performance of FedMobile-trained GAT is also dependent on the learning rates of training the expert and apprentice models. In general, we can select smaller learning rates to train the apprentice models than to train the expert model.

\section{Discussion}
This work has practical implications in privacy preserving machine learning for mobile health. First, FedMobile shows a promising solution for on-device deployment of Federated Learning to mitigate the memory scarcity issue in mobile devices. Additionally, the evaluation results imply that FedMobile can be used for privacy preserving individual level health monitoring and inference without compromising the predictive power of neural network models \cite{luo2022Demo}. Furthermore, at present, Centers for Disease Control and Prevention (CDC) relies on collaborative efforts from local clinics to state agencies to report influenza virus infection cases. However this case report system can be unpunctual and prone to underestimate. The limitations of traditional public health reporting and syndrome surveillance can be overcome by deploying our proposed mobile sensing solution in practice. Last but not least, interventions can be delivered to change individual's behavior (e.g., self-isolation) if influenza-like symptoms are detected. The interventions can be scaled up to a population level to combat the transmission of influenza virus. 

There are also several limitations in this work. First of all, although the mobile sensing data are collected in the wild, the experiments are conducted using simulations in an incremental learning setting using aggregated data from each client. In a real world situation, an individual user could be treated as a client, and data from each client can have heterogeneous distributions causing weight divergence problem, which may not have been appropriately addressed in our current proposed method. Secondly, in this study, we assume that the clients update their local models synchronously. However, in real world situation, each individual mobile device has different computational power and network access, making synchronous federated learning challenging.

In the future, we will deploy FedMobile on smartphones and develop a digital platform for decentralized human health monitoring. Furthermore, to achieve individual level health monitoring, it is critical to develop robust Federated Learning algorithm that can work on non-identically and independently distributed (non-IID) data with stationary global update. And we plan to enlarge the existing sensor network to include audio data, physiological data (i.e., heart rate), text messages \cite{zhan2021deepmtl}. Moreover, personalized FL for lifelong learning is another direction that is worth investigating.

\section{Conclusion}
In this work, we introduce FedMobile, an incremental semi-supervised Federated Learning framework for influenza-like symptom recognition using mobile sensing. FedMobile has demonstrated its potentials to train neural network models using continuously collected streaming data. It integrates Knowledge Distillation and Consistency Regularization, such that FL models can be updated in an online fashion to relieve computation and memory burden, and large amount of unlabeled data can be utilized to mitigate within-client inconsistency. The experiments we conducted using a real-world mobile sensing data generated from 800 participants demonstrate that FedMobile can recognize the existence of influenza-like symptoms in a privacy protected way without compromising the trained models' predictive capability \cite{qi2023blockchain}. 

\section*{Acknowledgement}
This work was supported by the DARPA Warfighter Analytics using Smartphones for Health (WASH) program.
.
\bibliographystyle{unsrt}
\bibliography{main.bib}

\end{document}